\documentclass[10pt,twocolumn,letterpaper]{article}

\usepackage{cvpr}              
\usepackage[accsupp]{axessibility}
\usepackage{graphicx}
\usepackage{amsmath}
\usepackage{amssymb}
\usepackage{booktabs}
\usepackage{enumitem}
\usepackage{float}
\usepackage{graphicx}

\usepackage[pagebackref,breaklinks,colorlinks]{hyperref}

\usepackage{color, colortbl}

\usepackage{soul}

\usepackage[capitalize]{cleveref}
\crefname{section}{Sec.}{Secs.}
\Crefname{section}{Section}{Sections}
\Crefname{table}{Table}{Tables}
\crefname{table}{Tab.}{Tabs.}

\def\cvprPaperID{11097} 
\def\confName{CVPR}
\def\confYear{2022}

\begin{document}

\title{SmartPortraits: Depth Powered Handheld Smartphone Dataset of Human Portraits for State Estimation, Reconstruction and Synthesis}

\author{Anastasiia Kornilova
\and
Marsel Faizullin
\and
Konstantin Pakulev
\and
Andrey Sadkov
\and
Denis Kukushkin
\and
Azat Akhmetyanov
\and
Timur Akhtyamov
\and
Hekmat Taherinejad
\and
Gonzalo Ferrer
\\ Center for AI Technology (CAIT), Skolkovo Institute of Science and Technology}

\twocolumn[{%
\renewcommand\twocolumn[1][]{#1}%

\maketitle
\begin{center}
    \centering
    \captionsetup{type=figure}
    \includegraphics[width=0.8\textwidth]{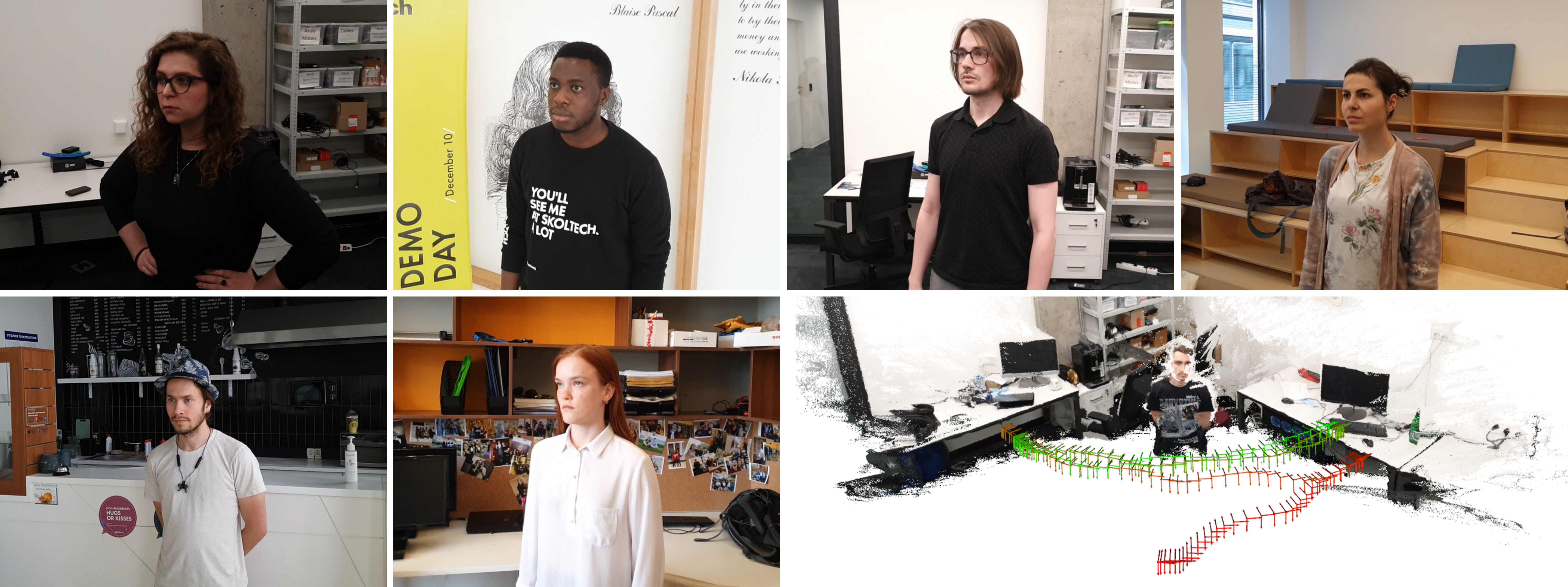}
    \captionof{figure}{From \textit{top-left:} examples of frames from the SmartPortraits dataset videos that capture human portraits in different natural environments, with varying lightning conditions, using a smartphone and external depth camera on a rig. \textit{Bottom-right:} recorded trajectory (red~-- initial time, green~-- end time) and dense reconstruction obtained by ACMP \cite{xu2020planar}.}
    \label{fig:teaser}
\end{center}%
}]

\begin{abstract}
We present a dataset of 1000 video sequences of human portraits recorded in real and uncontrolled conditions by using a handheld smartphone accompanied by an external high-quality depth camera.
The collected dataset contains 200 people captured in different poses and locations and its main purpose is to bridge the gap between raw measurements obtained from a smartphone and downstream  applications, such as state estimation, 3D  reconstruction, view synthesis, etc.
The sensors employed in data collection are the smartphone's camera and Inertial Measurement Unit (IMU), and an external Azure Kinect DK depth camera software synchronized with sub-millisecond precision to the smartphone system. During the recording, the smartphone flash is used to provide a periodic secondary source of lightning.
Accurate mask of the foremost person is provided as well as its impact on the camera alignment accuracy.


For evaluation purposes, we compare multiple state-of-the-art camera alignment methods by using a Motion Capture system.
We provide a smartphone visual-inertial benchmark for portrait capturing, where we report results for multiple methods
and motivate further use of the provided trajectories, available in the dataset, in view synthesis and 3D reconstruction tasks.
\end{abstract}


\section{Introduction}
\label{sec:intro}

Realistic rendering of people, and in general of objects, has recently achieved an unprecedented level of detail and realism \cite{mildenhall2020nerf, lombardi2019neuralvolumes, shysheya2019textured, gafni2021dynamic, zakharov2020fast, wang2021learning, Zakharov2019talking, aliev2020neural} with potential groundbreaking applications in telepresence, VR and AR.
Most of these methods have focused on static scenes and synthetic data, leaving apart the computational time required, which is still prohibitive.
In contrast, many potential usages of reconstruction and rendering are ideal candidate applications for smartphones, or other consumer-level devices, whose sensors are improving every year but still of limited quality. 
Our objective is to create a dataset that recreates {\em in the wild} conditions emulating smartphone users.

The SmartPortrait dataset\footnote{https://MobileRoboticsSkoltech.github.io/SmartPortraits/}  is an effort to bridge the gap between realistic raw data obtained from people, collected from a handheld smartphone and the down-stream reconstruction applications, for instance 3D portrait  reconstruction,  view  synthesis,  etc.
The key component that links both views is camera pose {\em state estimation}.
A usual practice is to obtain these poses by using a reliable but computationally demanding Structure from Motion (SfM) algorithm such as COLMAP \cite{schoenberger2016sfm} or multi-modal SLAM methods \cite{campos2021orb,openvslam2019,leuteneger2014okvis,Rosinol20icra-Kimera}.
The trajectories provided in our dataset emulate handheld movements, as if the user, or close-by users, were recording the sequences (see Fig.~\ref{fig:teaser}).

Many view synthesis methods \cite{mildenhall2020nerf,gao2020portrait,park2021nerfies,sevastopolsky2020relightable} generate their own datasets just by using state estimation methods. These single camera free-viewpoint images can only be considered if the scene is {\em static.}
We asked volunteers to stay as still as they could while recording them in a semicircular trajectory from close and mid distances.
We observed that most of the volunteers slightly changed their postures so we should expect some degree of displacement, which transforms the problem into {\em non-static}.

The SmartPortrait dataset is obtained in a variety of emplacements,  under different lightning conditions, plus a flashing light from the smartphone at regular intervals.
The smartphone camera is complemented with a high-quality depth sensor, adding robustness and multi-modality. 

We provide a recorded dataset consisting of smartphone video images,  IMU data, perfectly time aligned, and an external depth camera from Azure Kinect DK.
The evaluation includes two steps: first we compare the most promising methods with a reference trajectory obtained from a motion capture (MoCap) system. Second, for some environments, it is not possible to deploy the MoCap system. Therefore, we provide a reference trajectory, obtained from the previous best performing method, and provide an upper bound of the error by using a non-reference metric \cite{kornilova2021mom}.
In further evaluations, 
we benchmark multiple state of the art methods for visual SLAM, SfM and Visual-Inertial based methods.

Next, we want to connect the problem of camera pose estimation with two downstream tasks: 3D reconstruction with COLMAP~\cite{schoenberger2016mvs}, ACMP~\cite{xu2018monoperfcap} and SOTA view synthesis algorithms (NeRF \cite{mildenhall2020nerf}, FVS~\cite{Riegler2020FVS}, SVS~\cite{riegler2021stable}).
These applications will help us to understand the importance of pose estimation and its correlation with other tasks.





{\bf Ethical considerations.} We asked all participants in the dataset for a signed consent to record their portraits and publicly release them for  purely academic purposes. We explicitly indicated in the agreement their right, at any time, of removing all their data.
\section{Related work}
\label{sec:related}


Capturing human data is always addressed to a particular task, for instance, human faces, portraits, facial expressions, hand gestures, full bodies, etc. The common trait is that obtaining these data is a challenging process and different approaches exists to tackle them.
Our data include human portraits, or upper body of people, and there exists a relation to many other works on capturing human data. This section reviews the existing literature based on the sensor set used to obtain them. Later, we will discuss some applications and finally, we will present some state estimation methods as a requirement for single free-viewpoint recordings.

{\bf Motion Capture} systems \cite{vicon,optitrack} utilize multiple customized cameras to accurately detect reflective or infra-red markers. They are a popular method used to capture human data by tracking markers and synchronize them with video: HumanEva \cite{sigal2010humaneva}, Human3.6M \cite{ionescu2013human3} and INRIA \cite{Jinlong2016inria}. A negative effect is that it requires the volunteers to wear special suites over their bodies, changing their clothing appearance.

{\bf Multiple cameras} overcome this issue and remove the need for {\em markers}. They are a very popular method to capture body expressions and fine details, preserving visual appearance of the models.
Examples include
shape capture \cite{vlasic2009dynamic},
streamable free-viewpoint \cite{collet2015high},
AIST \cite{tsuchida2019aist},
Panoptic studio \cite{joo2017panoptic, joo2018total} with a mixture of  500 cameras, BUFF \cite{zhang2017detailed} for human pose and shape estimation, 
Humbi \cite{yu2020humbi} for body expressions, \cite{gao2020portrait, wang2021learning} for head portraits,
or photo-realistic full-body avatars \cite{bagautdinov2021driving}.
These settings are in practice very precise at capturing simultaneously the same event, e.g. a dynamic person. They are however expensive, difficult to deploy in different environments, and require a considerable amount of effort to calibrate and synchronize them.

Controlled {\bf lightning} conditions are also becoming an important feature for obtaining a fine detailed geometry reconstruction when digitizing humans \cite{vlasic2009dynamic,guo2019relightables, sevastopolsky2020relightable}.
SmartPortraits includes some images under smartphone flash conditions,
such that the  lightning source coincides with the optical sensor frame and creates a different outcome than under ambient lightning.

Some attempts have tried to lower the demanding requirements of multi-camera settings, where many sensors and lightning sources are required.
One solution is enhancing the data obtained with a single {\bf depth sensor} \cite{li20133d, shapiro2014rapid, bogo2015detailed, yu2018doublefusion, guo2017real} or multiple depth sensors \cite{zhang2012real, dou2016fusion4d, hong2018dynamic}.
Other approaches try to reduce the number of cameras in operation and still obtain reasonably accurate results \cite{zheng2021deepmulticap, zhang2021lightweight}.


At the extreme, one would desire a {\bf single camera} free-viewpoint, either taking multiple pictures or with a video \cite{xu2018monoperfcap, alldieck2019learning}. This is the aim of our dataset as well.

\begin{table*}
\fontsize{7.9}{7.2}\selectfont
\centering
\resizebox{\linewidth}{!}{ 
\bgroup
\def\arraystretch{1.5}%
\begin{tabular}{p{0.079\textwidth}p{0.136\textwidth}p{0.114\textwidth}p{0.114\textwidth}p{0.202\textwidth}p{0.114\textwidth}p{0.114\textwidth}p{0.128\textwidth}}
\toprule
 & EuRoC MAV & TUM-VI & TUM RGB-D & PennCOSYVIO & KAIST VIO & ADVIO & Ours \\
\midrule
Year & 2016 & 2018 & 2012 & 2016 & 2021 & 2018 & 2021  \\
Environ- ment & indoors & indoors/ \newline outdoors & indoors & indoors/ \newline outdoors & indoors & indoors/ \newline outdoors & indoors \\
Carrier & MAV & handheld & handheld/robot  & handheld & UAV & handheld & handheld \\

Focus & MAV VIO/SLAM & VIO & RGB-D SLAM & handheld VIO & UAV VIO & handheld VIO/SLAM & VIO/SLAM in\newline Human Digitiz.  \\

Cameras & 
stereo gray: 2x752x480 @20Hz &
stereo gray: 2x1024x1024 @20Hz &
RGB-D: \newline 640x480 @30Hz &
\textbullet\ 4 RGB: 1920x1080 @30Hz \newline
\textbullet\ stereo gray: 2x752x480 @20Hz \newline
\textbullet\ fisheye gray: 640x480 @30Hz &
\textbullet\ RGB: 640x480 @30Hz \newline
\textbullet\ stereo IR: 640x480 @30Hz&
\textbullet\ RGB: 1280×720 @60Hz \newline
\textbullet\ fisheye gray: 640x480 @60Hz &
\textbullet\ RGB: 1920x1080 @30Hz \newline
\textbullet\ depth: 640x576 @5Hz \\

IMUs &
ADIS16448 3-axis acc/gyro @200Hz &
BMI160 3-axis acc/gyro @200Hz &
Kinect 3-axis acc \newline @500Hz &
\textbullet\ ADIS16488 3-axis acc/gyro @200Hz \newline
\textbullet\ Tango 2 3-axis acc @128Hz \newline
\textbullet\ Tango 2 3-axis gyro @100Hz & 
Pixhawk 4 Mini 3-axis acc/gyro @100Hz &
MP67B 3-axis acc/gyro @100Hz &
\textbullet\ LSM6DSO 3-axis acc/gyro @500Hz \newline \textbullet\ MPU-9150 3-axis acc/gyro @500Hz\\

Time sync & hw & hw & hw & hw, sw & data & sw & hw, sw + frame sync \\
Point clouds & \checkmark (some seq) & $\times$ & \checkmark & $\times$ & $\times$ & \checkmark & \checkmark \\

Distance &
11 seq, 0.9 km &
28 seq, 20 km &
39 seq x several m &
4 seq, 0.6 km &
14 seq x several m &
23 seq, 4.5 km &
1000 seq, 6.6km \\
\midrule

Ground-truth & 
\textbullet\ 3D pos. (some seq), laser tracker @20Hz \newline
\textbullet\ 3D pose (some seq), MoCap @100Hz \newline
\textbullet\ 3D pcds (some seq), laser tracker & 
3D pose, MoCap @120Hz \newline (partial gt) & 
\textbullet\ 3D pose, MoCap @300Hz (partial gt) \newline
\textbullet\ 3D pcds, Kinect @5Hz & 
3D pose, visual markers @30Hz &
3D pose, MoCap @50Hz &
\textbullet\ 3D pose, IMU + manual position fixes @100Hz \newline
\textbullet\ 3D pcds, Tango @5Hz&
\textbullet\ 3D pose (some seq), MoCap @240Hz \newline
\textbullet\ 3D pose, COLMAP/RGB-D SLAM @5Hz \\

Acc. $\approx$ &
1 mm &
1 mm (static case) &
1 mm (relative) &
15 cm &
1 mm & 
0.1 - 1 m \cite{wang2019near} & 1 mm - 1 cm \\

\bottomrule
\end{tabular}
\egroup
} 
\caption{Overview of common Visual (V) and Visual Inertial (VI) benchmark datasets targeted at state estimation. 
}
\label{tab:related}
\label{tab:datasets_vo_vio_slam}
\end{table*}

From the perspective of datasets capturing human data for people reconstruction and rendering task, we observe the following modelling classes: full body modelling (DynamicFAUST~\cite{bogo2017dynamic}, BUFF~\cite{zhang2017detailed}, People Snapshot~\cite{alldieck2018video}), clothes modelling (3DPeople~\cite{pumarola20193dpeople}, SIZER~\cite{tiwari2020sizer}), \textit{head/torso portrait} modelling (UHDB11~\cite{toderici2013uhdb11}, Nerfies~\cite{park2021nerfies}, Portrait Neural Radiance~\cite{park2021nerfies}), or suitable for applications in couple of tasks (RenderPeople, Humbi~\cite{yu2020humbi}). There are also crowd-sourced datasets such as MannequinChallenge~\cite{li2019learning}, TikTok Dataset~\cite{jafarian2021learning} collected from social networks available for reconstruction tasks. Our dataset is unique since it records consumer-level data (smartphone) ``in the wild'' supported by high-quality external depth data.

Recently emerged neural implicit representation methods allow to bypass the need for obtaining accurate 3D structure of a scene and instead model it implicitly, e.g. by considering occupancy \cite{mescheder2019occupancy}, signed distance function \cite{park2019deepsdf} or volumetric density \cite{mildenhall2020nerf,  lombardi2019neuralvolumes}. In particular, there have been several works that successfully used neural implicit representation for creating realistic portrait avatars \cite{gao2020portrait, park2021nerfies, wang2021learning, gafni2021dynamic, sevastopolsky2020relightable}.



Unfortunately, when scenes include dynamic elements, and that is the case of video recording of people, the state estimation of camera poses or free-viewpoints and the 3D scene is not so trivial.
In the downstream tasks of 3D human reconstruction, two main variants exist:
free-form \cite{carranza2003free,saito2019pifu,saito2020pifuhd}
and model based \cite{loper2015smpl,alldieck2018video,omran2018neural}.


Accordingly, when reducing the number of cameras to a single one and working in {\em non-static} conditions because the volunteers in our dataset stand still but not immobile, then,
{\bf state estimation} becomes the key ingredient that allows a handheld single camera video to be used for human digitization.
Either if camera poses are compensated while learning a model \cite{lin2021barf} or estimated as recorded, the quality of these is going to be determinant for any downstream task.

To the best of our knowledge, there are no datasets that directly address the evaluation of state estimation approaches when a human is in the main focus of sensors.
State estimation of camera poses include techniques such as 
Visual Odometry (VO) \cite{Forster2014ICRA,DBLP:journals/corr/EngelKC16},
Visual-Inertial Odometry (VIO) \cite{leuteneger2014okvis,qin2019local,PRCV-LiYHZB2019,bloesch2017rovio}. Variants of Simultaneous Localization and Mapping (SLAM), which include a loop in estimation for global pose alignment, either 
Visual SLAM  (V-SLAM) 
\cite{mur2015orb,gao2018ldso,mur2017orb,openvslam2019} or 
Visual-Inertial SLAM (VI-SLAM)
\cite{qin2017vins,Rosinol20icra-Kimera,Geneva2020ICRA,campos2021orb} and Structure from Motion (SfM) \cite{schoenberger2016sfm}, all of them relevant to the be applied to the sensor data available in a smartphone.
To date, there are numerous datasets available \cite{Burri25012016, jeon2021run, tumvi, sturm12iros, penncos, Cortes_2018_ECCV, geiger2013vision, Delmerico2019drone, carlevaris2016university, ceriani2009rawseeds, zuniga2020vi, wang2019near} that vary greatly by their focus, recording environment, sensor carriers as well as by the amount of data recorded and the accuracy of the ground truth.
We briefly describe the main features of the main datasets and give a comparison with our dataset (see Table \ref{tab:related}).

The EuRoC Micro Aerial Vehicles (MAV) dataset~\cite{Burri25012016} focuses on VIO and SLAM for MAVs as well as 3D reconstruction. 
The authors employ a stereo pair of cameras hardware-synchronized with an IMU installed on a MAV for acquiring the data sequences in two indoor environments. 
Kaist VIO \cite{jeon2021run} is another indoor dataset that focuses on VIO for aerial vehicles, it specifically addresses challenging scenarios for VIO that contain pure rotational/harsh motions. 
TUM-VI \cite{tumvi} is a dataset for the purpose of evaluating VIO algorithms. Compared to other mentioned datasets it stands out by its size, diversity of the recorded sequences, as well as uses higher resolution cameras. 
TUM RGB-D \cite{sturm12iros} features only indoor sequences captured with a Kinect sensor that is handheld or mounted on a robotic platform. The dataset includes challenging scenarios for the proper evaluation of RGB-D SLAM  approaches. 
PennCOSYVIO \cite{penncos} is one more VIO benchmark that contains diverse challenging sequences. It includes not only rotational motions but hard visual conditions as well. The dataset was captured using a larger number of sensors than any other related datasets: 3 GoPro cameras, an integrated VI-sensor and 2 Google Project Tango tablets mounted on a rig. However, as pointed out in \cite{tumvi} the setup yields lower synchronization accuracy between cameras and IMUs when compared to datasets like TUM-VI. 
ADVIO \cite{Cortes_2018_ECCV} focuses on benchmark VIO and SLAM methods for smartphones and mobile devices with low-cost sensors. It contains different large indoor and outdoor environments recorded in public places. 

Our dataset contains recordings of people in indoor environments and focuses on the diversity of people, their clothing, the environments and the lightning conditions, totally uncontrolled aiming to recreate every-day life conditions of smartphone users.
Compared to PennCOSYVIO, KAIST VIO and ADVIO that use heterogeneous sensors too, we employ a significantly more precise hybrid hw/sw synchronization technique from \cite{faizullin2021synchronized} (see Sec.~\ref{sec:sys_sync}).
Due to recordings surroundings peculiarities we provide pseudo-ground truth poses based on the gathered sensors data like in \cite{Cortes_2018_ECCV} the quality of whose is estimated in a manner similar to \cite{penncos} by capturing verification sequences (see Sec.~\ref{sec:gt}).


\section{Recording Platform}
\label{sec:sys}

\begin{figure}[t]
  \centering
  \includegraphics[width=0.99\linewidth]{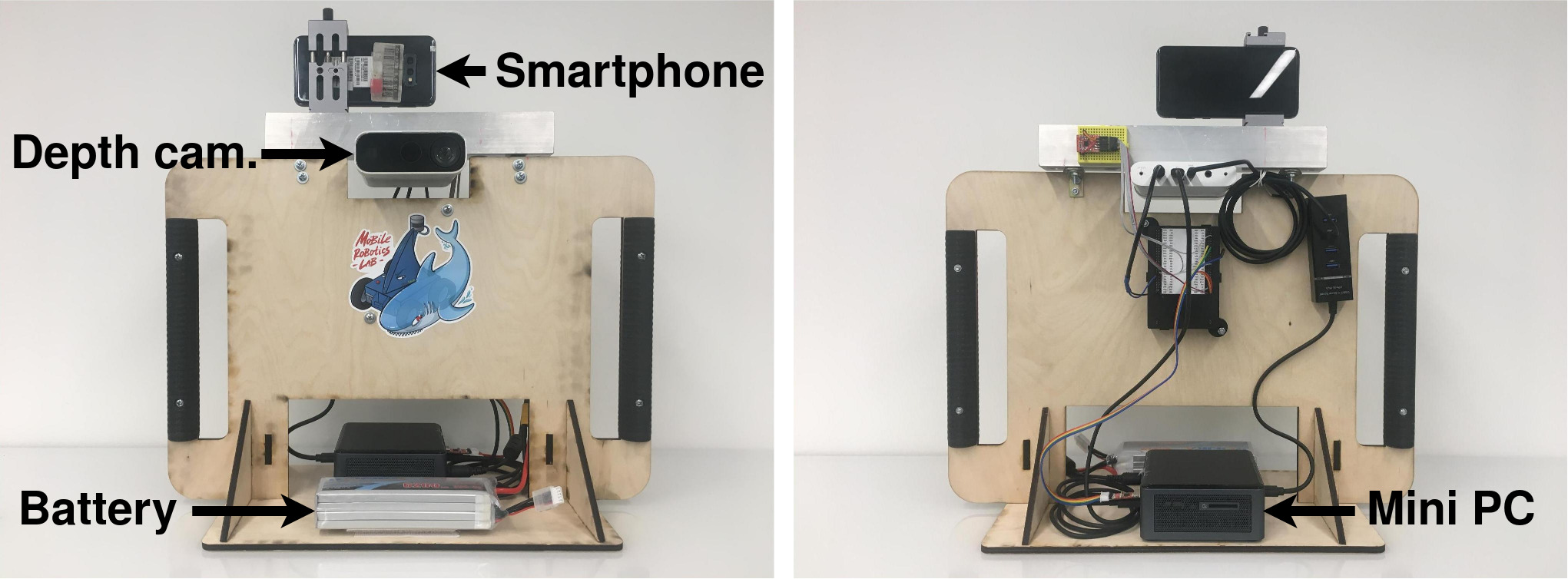}
  \caption{Front and back view of the recording platform. }
  \label{fig:system_common_view}
\end{figure}

Our dataset aims to provide human portrait data in the realistic environments captured by a middle-price smartphone. To meet these requirements, we have designed a portable handheld platform with a \textbf{Samsung S10e} smartphone (RGB camera, 1920x1080 p, 30 fps; IMU, 500 Hz, flasher, 1 Hz) and a high-end depth sensor \textbf{Azure Kinect DK} (depth camera, 640x576 p, 5 fps).  The high-quality external depth camera is chosen instead of the smartphone with a built-in sensor as (i) the modern smartphone depth images are still not as high-quality as external depth sensors, and (ii) it is not possible to record RGB and depth tracks by the smartphone on high frequency simultaneously. A common view of the system is presented in Fig.~\ref{fig:system_common_view}. 


Specifics of our recording case~--- dynamic camera movements close to real-life handheld capture and a person in foreground with non-stationary pose (blinking eyes, small movements of the person because of breath, coordination, heart beats). 


\subsection{Time and Frame Synchronization}
\label{sec:sys_sync}

The independence of smartphone and depth camera adds an additional challenge to the time synchronization. If frames from both sensors are captured at slightly different instants of time, several tens of $ms$, this degrades the quality of the camera pose estimation.

To synchronize the cameras, we introduce a two-step sync process.
First, the time domains between two sensors are synchronized by Twist-n-Sync algorithm~\cite{faizullin2021twist} which is not affected by network asymmetry, in contrast to network-based protocols like NTP. And in the second step~-- the synchronization of frame capturing moments from both sensors is done. For solving that, the grabbing of the smartphone's framing phase is performed via remote API interface. Then depth camera triggering is automatically tuned to this phase as explained in~\cite{faizullin2021synchronized}.
The sync used provides sub-millisecond accuracy.

\subsection{Calibration}
\label{sec:sys_calib}

The full intrinsic and extrinsic calibration of smartphone camera and smartphone IMU, is obtained by the Kalibr toolbox~\cite{rehder2016extending} with 6x6 AprilGrid array of 3x3 cm AprilTags~\cite{olson2011apriltag} as the visual markers. 

\begin{figure}[t]
  \centering
  \includegraphics[width=0.99\linewidth]{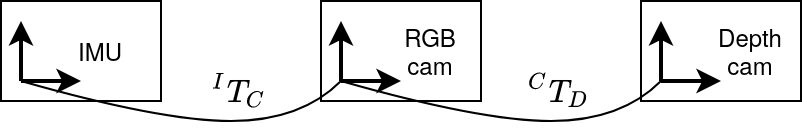}
  \caption{Obtained relative transformations by calibration. $^I T_C$~-- smartphone camera in smartphone IMU reference frame, $^C T_D$~-- depth camera in smartphone camera frame. $^C T_D$ is found as $^C T_D = ^C T_R \cdot ^R T_D$, where $^C T_R$ is Azure RGB camera in smartphone camera frame obtained by Kalibr, $^R T_D$ is factory-known Azure RGB camera in Azure depth camera frame. Azure RGB is only used for this procedure.}
  \label{fig:transforms}
\end{figure}

To find depth-to-smartphone camera transformation, firstly, we obtained the Azure RGB to smartphone camera transform. Then, combined it with the factory-known Azure depth (infra-red) to Azure RGB camera transform. This method gives much better accuracy than direct depth-to-smartphone camera transformation with low-quality infra-red camera. We use Azure RGB only for this procedure. All the obtained transformations are shown in~Fig.\ref{fig:transforms}.

Smartphone camera is rolling-shutter type; however, we applied global-shutter camera model during calibration to feed the calibration parameters correctly to the methods we compared in Sec.~\ref{sec:eval}. 
Standalone IMU calibration is also performed. IMUs noise parameters were borrowed from~\cite{lsm6dso} and~\cite{nikolic2016non} for smartphone and standalone IMU respectively.


\label{sec:calib_cam2depth}

\section{Dataset}

Our dataset contains 1000 records of 200 people, with natural clothing, captured in different native locations and poses. Every record consists of synchronized smartphone data (Full HD RGB video, flash timestamps, timestamped accelerometer and gyroscope measurements) and the depth images from external high-quality depth sensor. The dataset also contains reference ground truth trajectories of the smartphone camera obtained as described in Sec.~\ref{sec:gt} and segmentation masks for the person.
We supplement the dataset by labels for genders and difficult cases for rendering like volume hairs/beard/glasses.
Data parameters and sampling rates of the sensors are presented in Table.~\ref{tab:datasets_vo_vio_slam}.

\subsection{Collection Process and Statistics}

During every recording process, three people are involved: 
(1) a volunteer who is being filmed, (2) an operator that carries the recording platform, and (3) an assistant that monitors the correctness of the recording through SSH.
The volunteer is asked to stand or sit still. The operator carries a recording platform around the person at the subject's face height as depicted in Fig.~\ref{fig:teaser}. 
The recording trajectory begins in front of the person to capture the whole scene, then the operator moves to a side of the volunteer and makes four 100-120 degrees circular arcs around the model. The entire trajectory is shown in Fig.~\ref{fig:teaser}. 
The timestamps of every arc edge are marked online during recording by the assistant in an automated manner. In post-processing  stage, the whole trajectory could be split into separate arcs applying these marks.


Every person is captured in 5 different poses~--- 3 in a standing position (straight, hand on hips, with head turned) and 2 in a sitting position (straight, with head turned). 
Standing and sitting positions were captured from a distance of about 2 and 1 m respectively. 
During the recording, blinking flash on the smartphone is turned on with a frequency of 1 Hz to relight the model. Effect of re-lightning is more distinguishable on sitting positions since they are captured from a closer distance.


Data collection is performed in 5 different locations of native indoor environments: cafeteria, lab, office, campus entrance, and student council. Their common view is demonstrated in Fig.~\ref{fig:teaser}. 
The average length of the trajectories for staying and sitting position are 7.14 and 5.8 m accordingly. The total duration of all tracks is 11 hours and 6 minutes, and the total length is 6610 meters.

\begin{figure}[t]
  \centering
   \includegraphics[width=0.95\linewidth]{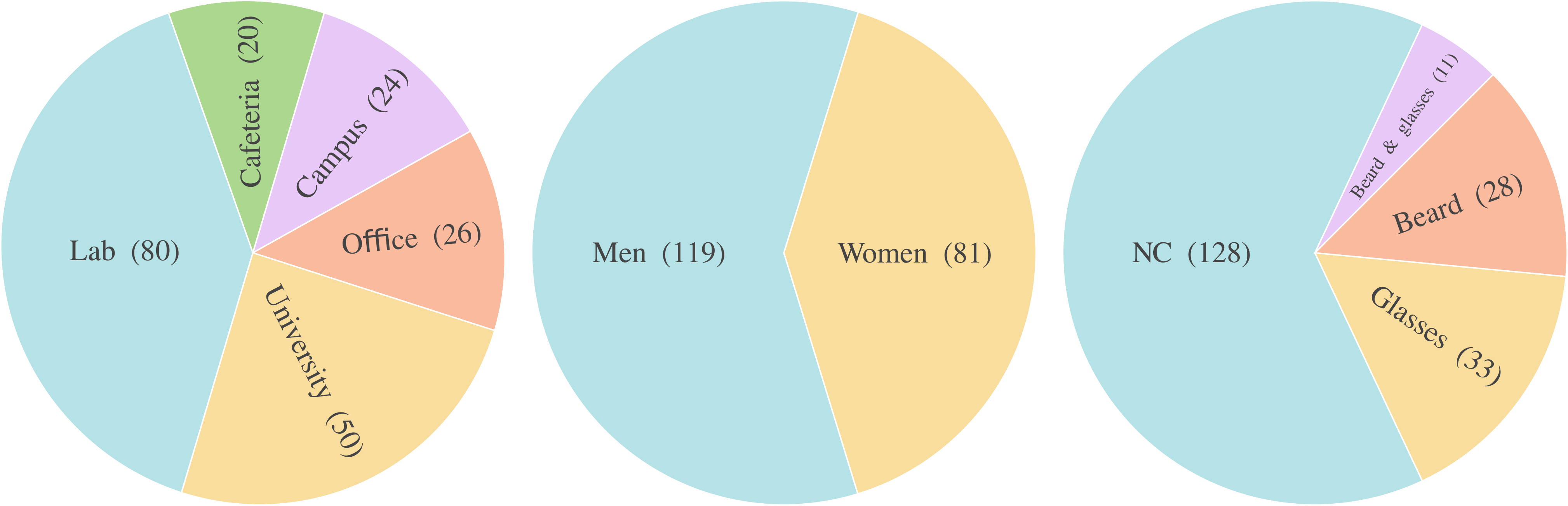}
  \caption{Dataset statistics. {\em Left:} Locations of recording. 
  {\em Center:} Gender {\em Right:} Appearance.}
  \label{fig:dataset_stat}
\end{figure}

SmartPortrait contains people of different gender, appearance, clothing, hairstyles, etc. Statistics are shown in Fig.~\ref{fig:dataset_stat}.


\subsection{Segmentation Masks}

Along with recorded data, we also provide segmentation masks of humans on the images. This information could be used for filtering out potentially dynamic landmarks of the scene on the trajectory estimation step (blinking eyes, subject movements) as demonstrated in Sec.~\ref{sec:gt} or for separation of portraits parts from the scene for only-person 3D reconstruction.
For this task, we design a semi-automated labeling process, based on U2-Net~\cite{qin2020u2} that is pre-trained on people masks from the Supervisely Person Dataset~\cite{supervisely}. Usage of this method on our data overestimates the person mask, also covering some parts of the background. DBSCAN clusters the masked part by using the depth component, discarding scene parts that are not related to the foreground. Finally, segmentation results are assessed visually by labelers.

\section{Evaluation}
\label{sec:eval}

\begin{figure*}[t]
  \centering
  \includegraphics[width=0.95\linewidth]{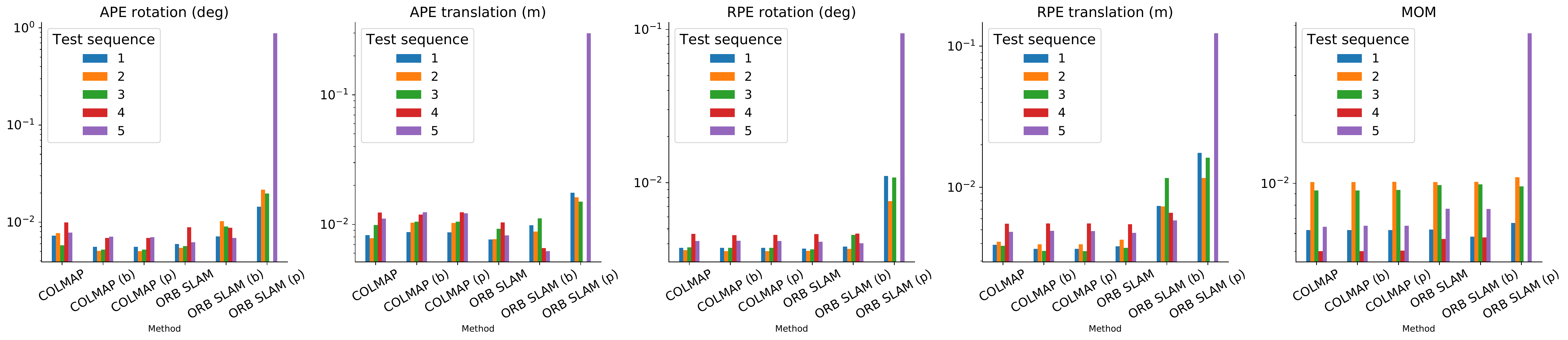}
  \caption{Full-reference (APE/RPE) and no-reference (MOM) metric statistics for COLMAP and ORB SLAM (RGB-D) for 5 test sequences with MoCap ground truth poses. (b) and (p) indicate that only background keypoints and person keypoints correspondingly were considered for pose estimation. One pose for ORB SLAM (p) is not converged, therefore its values are excluded from the evaluation.}
  \label{fig:mocap_eval}
\end{figure*}

\begin{figure*}[t]
  \centering
  \includegraphics[width=0.9\linewidth]{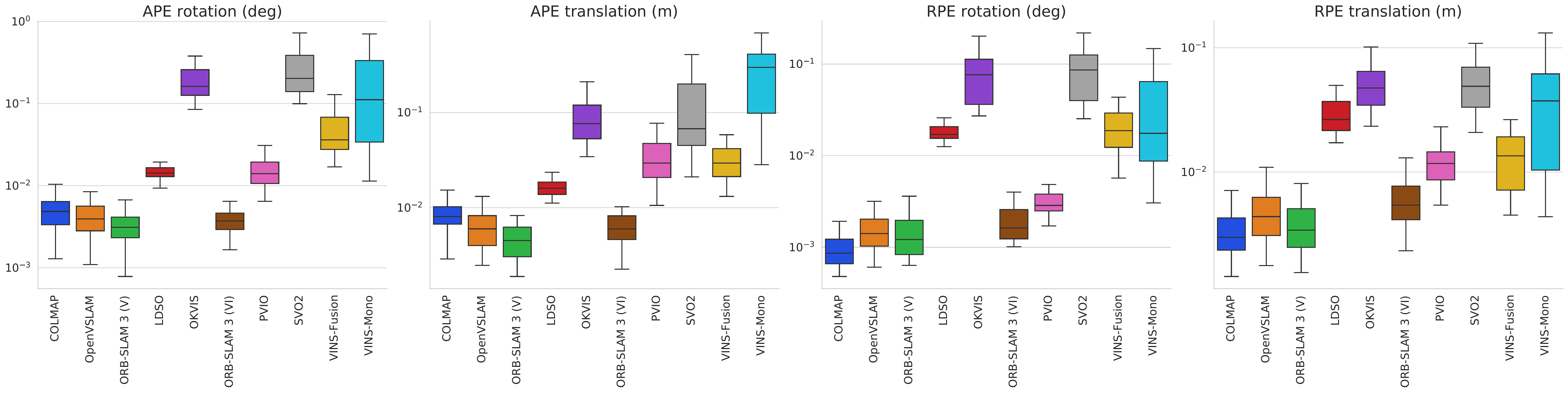}
  \caption{Evaluation of V/VI methods that employ only smartphone data (frames and IMU) on benchmark sequences.}
  \label{fig:vi-eval}
\end{figure*}

The evaluation part tackles two main questions~--- (1) how to find the best way of calculating pseudo-ground truth poses for our dataset and (2) investigate the performance of V and VI state estimation methods on smartphone data only.
V denotes all visual methods: VO and V-SLAM; the same applies for VI.

\subsection{Metrics}

\textbf{Full-reference metrics.} Among the class of full-reference metrics where the reference trajectory (ground truth) is available, we consider RMSE statistics on Absolute Pose Error (APE) and Relative Pose Error (RPE) for the rotation and translation parts. In particular, for translation APE, we apply the Umeyama alignment~\cite{arun1987least,umeyama1991} between a pair of trajectories if expressed in different origin frames. For rotation APE, the Umeyama alignment is followed by the trajectory's reference frame transformation.

\textbf{No-reference metrics.} No-reference metrics are alternative to the full-reference metrics when the reference trajectory is not available or its quality is disputable. In our work, we use Mutually Orthogonal Metric (MOM)~\cite{kornilova2021mom} that measures quality of the trajectory by evaluating quality of the map aggregated from point clouds registered via the trajectory poses. MOM provides stronger correlation with RPE error in comparison to its competitors~\cite{razlaw2015evaluation}. In our setting, MOM uses the point clouds converted from depth images.

In order to apply MOM on trajectories ambiguous to scale (e.g., COLMAP), we optimize the scale factor w.r.t. MOM metric~---  which assumes that the correct value of scale is reached at the optimum in the metric, when the aggregated map of point clouds is at its best condition.

\subsection{Ground Truth Trajectories}
\label{sec:gt}

The majority of the dataset sequences (see Fig. \ref{fig:dataset_stat}) are captured in public places or areas where either applying conventional methods of acquiring ground-truth poses (e.g., MoCap) are not feasible, or such methods disrupt the nativity of the surroundings (e.g., visual markers).

Therefore, it is required a no-reference method in order to validate the obtained trajectories when MoCap data is not available. Below, we will present a procedure to select a new reference trajectory and an upper bound of its error.


\textbf{Methods.} Since the dataset is targeted on both state estimation and reconstruction/synthesis domains, we consider the main methods typically used by the community that make use of the sensors: RGB, depth cameras and IMU.
From \textit{reconstruction and rendering} field experience, we consider the COLMAP \cite{schoenberger2016sfm} Structure-from-Motion (SfM) pipeline that is de-facto standard tool in this area and usually employed as ground truth. COLMAP uses only RGB data and therefore its trajectory is defined up to a scale factor, that limits its usage in state estimation tasks. In addition, we consider the class of RGB-D SLAM algorithms that are able to provide poses and scale is observable.
Based on the wide evaluation of RGB-D SLAM methods done in~\cite{zhang2021survey} we choose ORB-SLAM (RGB-D) \cite{mur2017orb}, implementation from \cite{campos2021orb}, as one with the lowest trajectory error.

\textbf{MoCap Test Sequences.} To assess the accuracy of the ground truth poses, we record several testing sequences in the laboratory environment where the use of a more accurate ground truth acquisition method is possible. In particular, we utilize OptiTrack MoCap system \cite{optitrack} to record 5 testing sequences of one person in the common dataset format. MoCap is synchronized with the platform offline by the Twist-n-Sync algorithm~\cite{faizullin2021twist}. 
The extrinsic parameters calibration requires to calculate:

\begin{equation}
    \min_{X,Y\in SE(3)} \sum_{i} || \log \big( Y \cdot T_M(i) \cdot X \cdot T_C^{-1}(i)\big) ||,
\end{equation}
where $T_M(i)$ is the trajectory given by the MoCap at time $i$, $T_C$ is pose at the camera frame calculated by the algorithm, $X$ is the transformation between the camera optical center and the tracked object in the MoCap and $Y$ is the transformation between the origin frame of the MoCap and the origin of the SLAM algorithm (usually first frame).

\textbf{Results.} In order to support the selection of the pseudo ground truth, we evaluate COLMAP and ORB-SLAM (RGB-D) (actually, virtual stereo) on the MoCap test sequences by using the described above full-reference and no-reference metrics. Because the landmarks on the volunteer body might be non-static (person can breath, blink), we consider three modifications of COLMAP and ORB-SLAM (RGB-D) -- using the whole scene, using mask for background, using mask for person. The evaluation results are presented in the Fig.~\ref{fig:mocap_eval}. Both methods demonstrate almost the same performance on the considered metrics, being close to the resolution limit of the MoCap system. The usage of masks does not affect the performance in case of COLMAP that could be explained by photometric consistency imposed by the algorithm. ORB SLAM performs worse when only part of the scene is visible. 

\begin{figure}[t]
  \centering
  \includegraphics[width=0.9\linewidth]{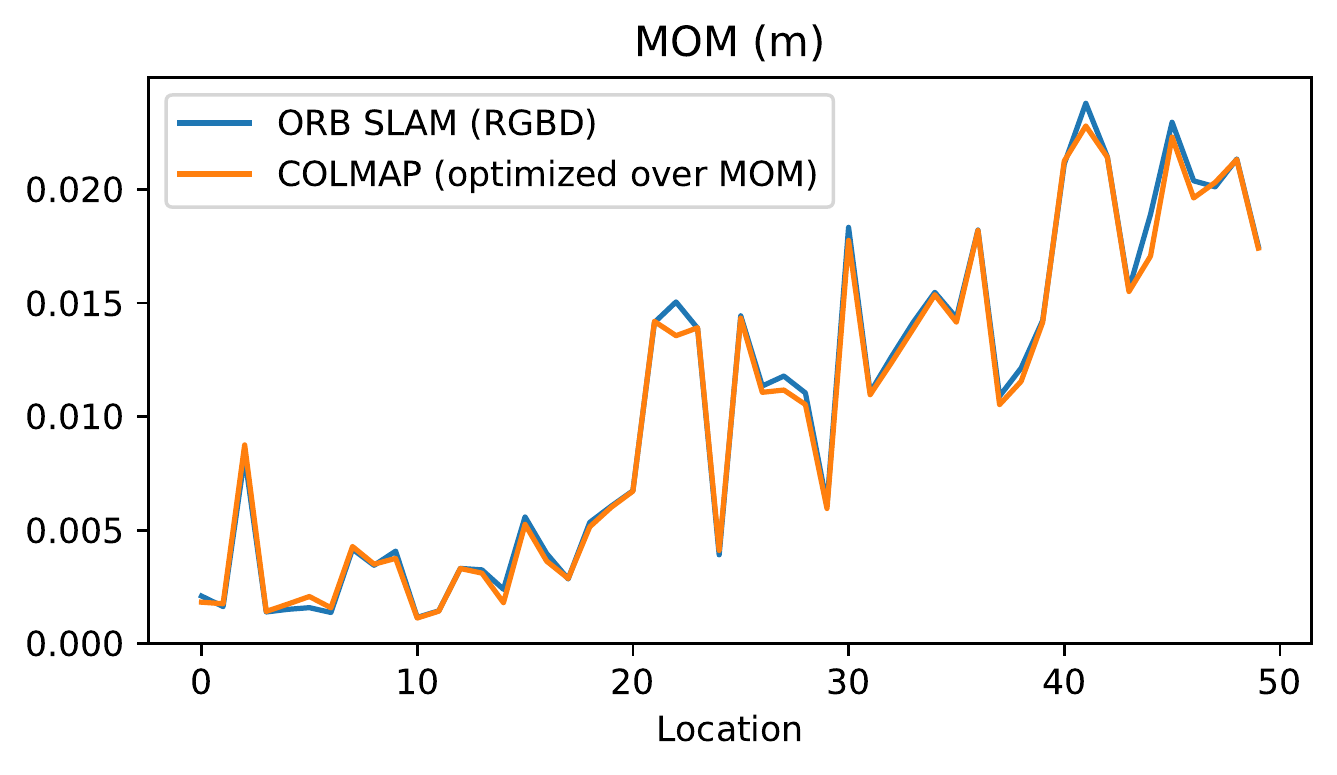}
  \caption{MOM generalization to 50 other scenes from the dataset.}
  \label{fig:mom_generalization}
\end{figure}

\textbf{Eval Generalization.} The above evaluation on the MoCap test sequences with ground truth available is limited only to the  lab location. To extend it to all locations that our dataset covers, we must consider the comparison of COLMAP and ORB-SLAM (RGB-D) by  using the no-reference metric MOM. For that, we select 10 trajectories from every location that result into 50 test sequences. Evaluation performance is presented in Fig.~\ref{fig:mom_generalization} and it allows to make the following two conclusions. Firstly, since MOM measures the dispersion of deviations on planar surfaces in the aggregated map it can be noticed that for the majority of locations it does not overcome 2 cm. This value is comparable to the depth sensor noise, that means that both methods give relatively good trajectories from the state estimation point of view. Secondly, COLMAP performs slightly better than ORB-SLAM (RGB-D), although it requires post-processing for revealing the scale. We will provide the obtained trajectories of both methods as the pseudo-ground-truth methods.

\begin{figure*}[t]
  \centering
  \includegraphics[width=0.9\linewidth]{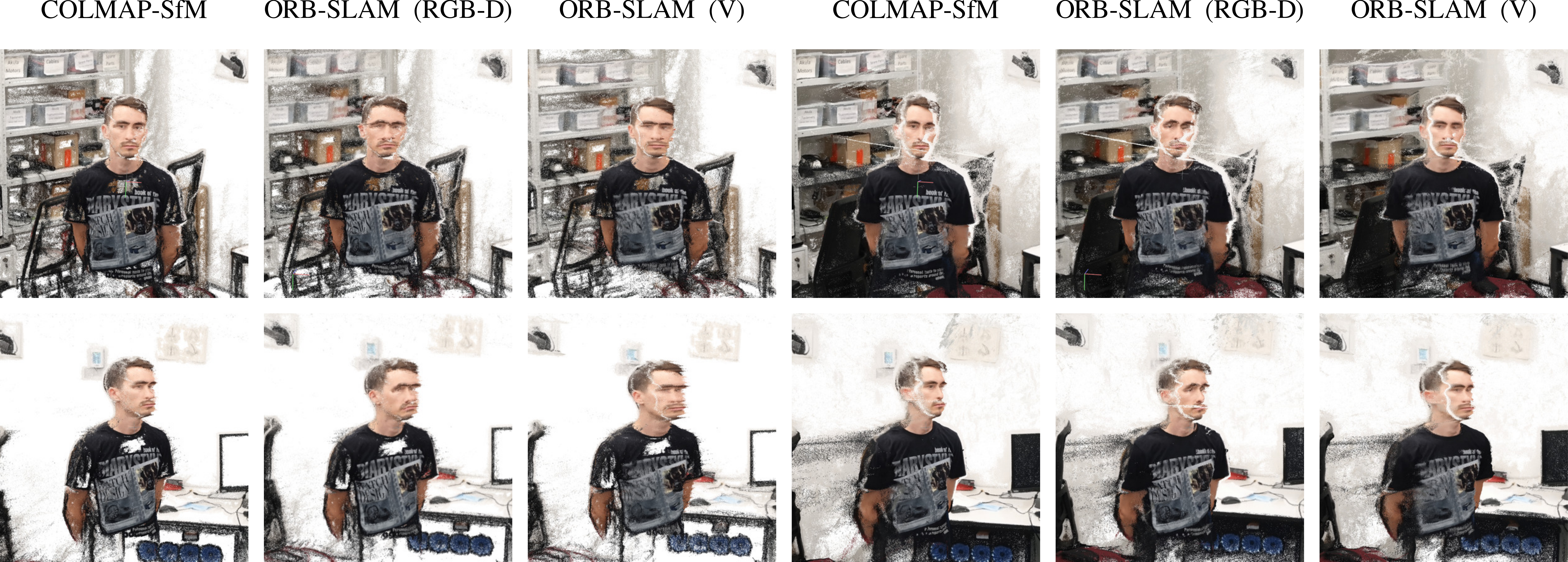}
  \caption{Qualitative demonstration of dense reconstruction from different views using COLMAP-MVS (3 left columns) and ACMP (3 right columns) on poses from COLMAP-SfM, ORB-SLAM (RGB-D), ORB-SLAM (V).}
  \label{fig:ColmapACMP}
\end{figure*}

\subsection{V and VI Evaluation}

One of the motivations of our work is the study the potential of applications of V/VI methods using smartphone-only data targeted to the domain of human portraits. In this section, we provide evaluation of different state-of-the-art methods and a baseline for future comparisons. In addition, to all the considered methods, we deliver configuration and calibration files for methods to be used with our data for benchmark.

\textbf{Methods.} In evaluation we consider two classes of methods: Visual (V) and Visual Inertial  (VI) methods. Considering recent exhaustive evaluations~\cite{jeon2021run, delmerico2018benchmark}, we order top-rated V/VI methods. The considered methods for both classes are: for V~--- OpenVSLAM\cite{openvslam2019}, ORB SLAM Monocular\cite{campos2021orb}, LDSO\cite{gao2018ldso} and
COLMAP \cite{schoenberger2016sfm}.
For VI (ordered by performance on other datasets)~-- ORB-SLAM 3 (VI) \cite{campos2021orb}, Kimera VIO\cite{Rosinol20icra-Kimera}, OpenVINS\cite{Geneva2020ICRA}, VINS-Fusion\cite{qin2019local}\cite{qin2019global}, VINS-Mono\cite{qin2017vins}, PVIO\cite{PRCV-LiYHZB2019}, SVO.2\cite{7782863}, MSCKF\cite{4209642}, OKVIS\cite{leuteneger2014okvis}, ROVIO\cite{bloesch2017rovio}. Some of the methods (Kimera VIO, OpenVINS, MSCKF, ROVIO) were discarded because they require recording device to be static over the first seconds of the trajectory for initialization, whereas our use case does not cover such scenario.

\textbf{Benchmark Dataset} To evaluate the performance of the methods, we uniformly pick 2 sequences for every combination of location and volunteer pose, resulting in a total of 50 evaluation sequences. As ground truth, we consider the trajectories produced by ORB-SLAM (RGB-D) that, as demonstrated in Sec.~\ref{sec:gt}, provides excellent performance.


\textbf{Results.} The evaluation results on the set of the full-reference metrics are presented in Fig.~\ref{fig:vi-eval}. Because the pseudo GT provides a statistical bound, it could be used for exact ordering in cases when the order of the error magnitude is higher than the error between MoCap and pseudo GT. In particular, in our comparison we can order only the next V/VI methods: LDSO, OKVIS, PVIO, SVO2, VINS. In general, we can observe that V methods perform more accurately in rotation and translation than VIO methods. The VIO method's accuracy varies, where ORB 3 VI performing as accurate as V methods. All VI have better than 1 degree of accuracy in absolute rotation error.
\section{Application}
\label{sec:application}

\textbf{3D reconstruction.} For qualitative comparison on 3D reconstruction, we provide poses obtained from the top performing methods for state estimation from every class and use two state-of-the-art methods for 3D reconstruction: COLMAP multi-view stereo (MVS)~\cite{schoenberger2016mvs} and ACMP~\cite{xu2020planar}. The demonstration of the obtained 3D scenes is presented in Fig.~\ref{fig:ColmapACMP} from two views~--- center and the edge of the trajectory arcs. For both reconstruction pipelines, COLMAP trajectory produces less distorted reconstruction, overcoming ORB-SLAM (RGB-D) and (V) versions. It could also be noticed that from the trajectory border point of view 3D reconstruction has less quality.
The solution from the VI method produced an incorrect reconstruction, perhaps due to its error.

\begin{figure}[t]
  \centering
  \includegraphics[width=0.85\linewidth]{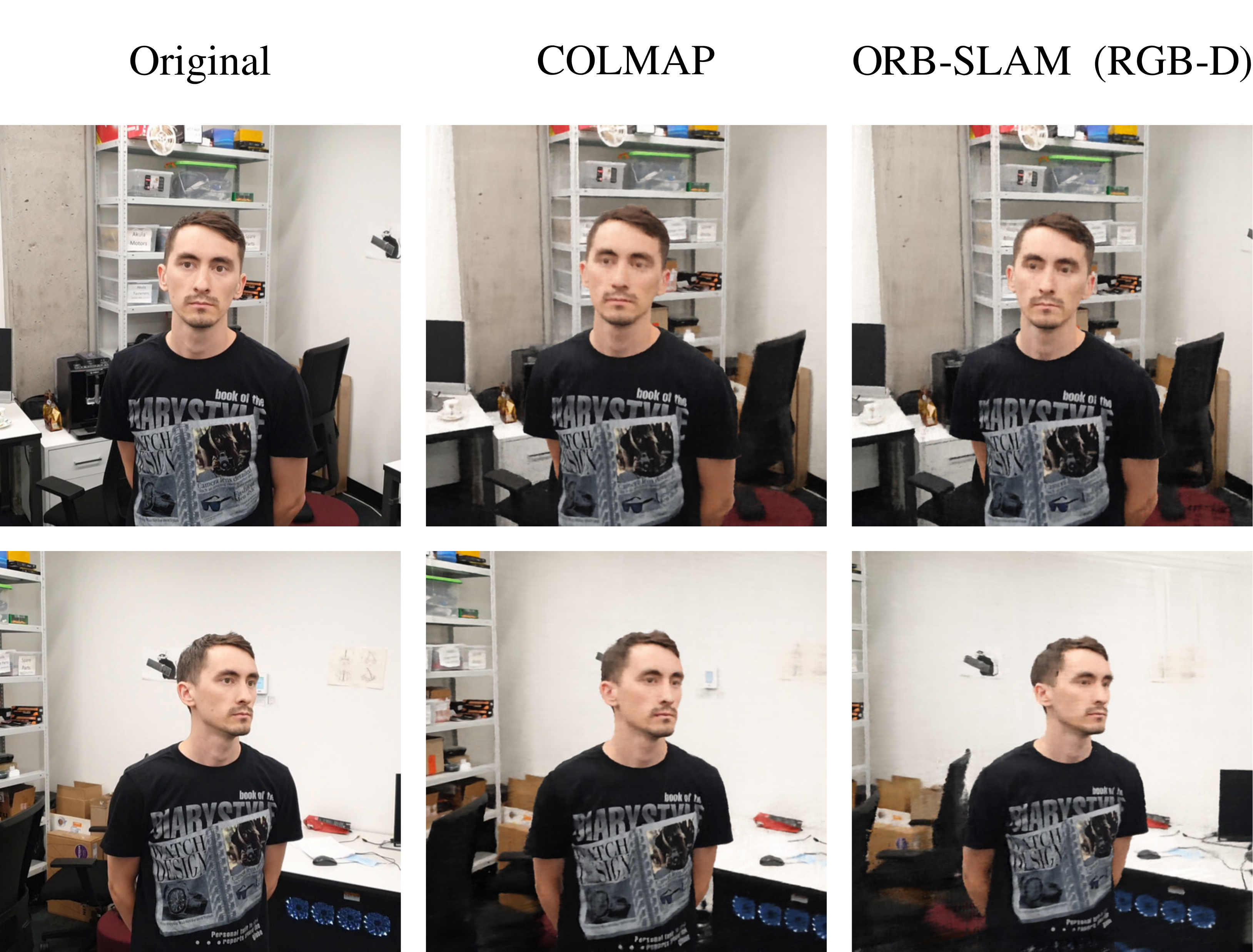}
  \caption{Qualitative demonstration of the NeRF novel-view synthesis algorithm on new poses, not observed before.}
  \label{fig:nerf}
\end{figure}

\begin{figure}[t]
  \centering
  \includegraphics[width=0.9\linewidth]{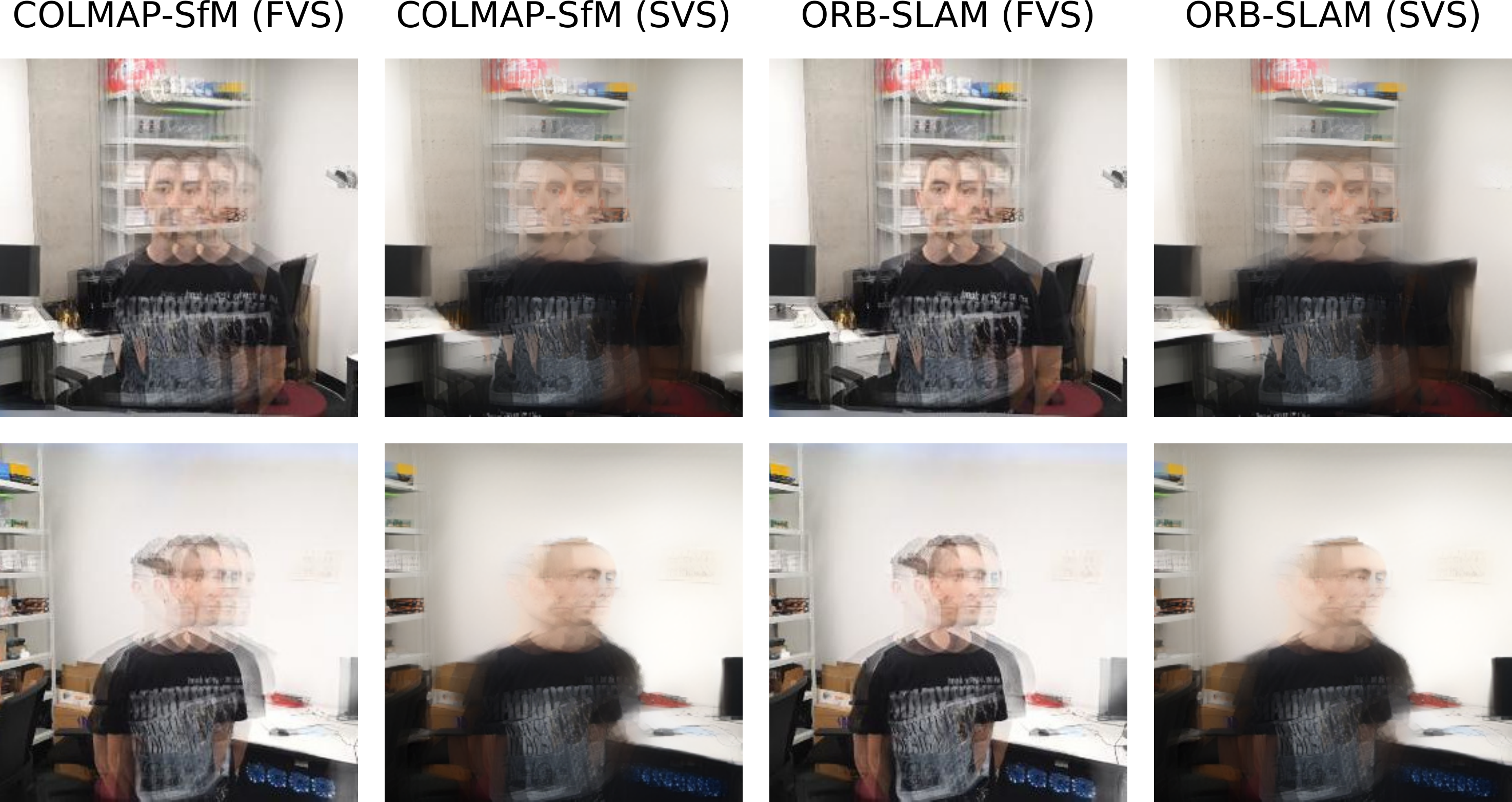}
  \caption{Qualitative demonstration of the FVS and SVS novel-view synthesis algorithms on new poses, not observed before.}
  \label{fig:fvs_nvs}
\end{figure}

\textbf{View Synthesis.} We provide a qualitative comparison of the considered VO-SLAM methods evaluated on an image synthesis problem. For that we considered SOTA methods: Neural Radiance Field \cite{mildenhall2020nerf} (NeRF), and generalized NeRF approaches~--- FVS~\cite{Riegler2020FVS} and SVS~\cite{riegler2021stable} in their default pre-trained versions. The data provided to methods is the solution of the trajectories. For NeRF algorithm, that is optimized per each scene, COLMAP provides the best qualitative results, the objects are coherently synthesized, a little blurred. ORB-SLAM (RGB-D) result shows some inconsistencies on the rendering (Fig.~\ref{fig:nerf}-{\em Bottom-right}) and overall less definition than COLMAP, although they both showed a similar trajectory error. Results of FVS and SVS are presented in Fig.~\ref{fig:fvs_nvs}. For both, COLMAP and ORB,  provide less quality than NeRF. It could be explained by the difference in poses configuration w.r.t original methods data on which methods were trained.

\section{Discussion}

One of the questions that this paper raised in the introduction is: {\em Are we ready to calculate handled trajectories in the wild and convert them into 3D portraits of people?}


Real time VI methods do not perform as accurately as V methods (some of them not real-time). Still, the accuracy achieved is remarkable, providing very accurate trajectories; probably, they could be better if IMUs were properly initialized.

Despite 
their accuracy, the results obtained on the applications (NeRF and reconstruction) could be improved.
This question is to be explored, but our qualitative results hint that trajectory error is not perfectly correlated with the downstream task, either for synthesis or for reconstruction. An explanation can be the photo-metric consistency, which is more important than the trajectory error.
A corollary of this: perhaps the solution is not to jointly optimize trajectories and maps, but simply optimize the map, allowing for small disturbances on the camera poses from a reasonable initial solution.
We plan to further investigate this issue. 

{\bf Potential Negative Societal Impact.}
Realistic human data are required to achieve a future of immersive VR and tele-presence. However, there are other potentially dangerous uses of these technology such as identity theft or fake news.

{\bf Acknowledgments.} This research is based on the work supported by Samsung Research, Samsung Electronics.

{\small
\bibliographystyle{ieee_fullname}
\bibliography{egbib}

\begin{thebibliography}{100}\itemsep=-1pt

\bibitem{optitrack}
{OptiTrack}.
\newblock \url{www.optitrack.com/}.

\bibitem{supervisely}
Supervisely person dataset.
\newblock
  \url{https://supervise.ly/explore/projects/supervisely-person-dataset-23304/datasets}.

\bibitem{vicon}
Vicon motion systems.
\newblock \url{www.vicon.com}.

\bibitem{aliev2020neural}
Kara-Ali Aliev, Artem Sevastopolsky, Maria Kolos, Dmitry Ulyanov, and Victor
  Lempitsky.
\newblock Neural point-based graphics.
\newblock In {\em Computer Vision--ECCV 2020: 16th European Conference,
  Glasgow, UK, August 23--28, 2020, Proceedings, Part XXII 16}, pages 696--712.
  Springer, 2020.

\bibitem{alldieck2019learning}
Thiemo Alldieck, Marcus Magnor, Bharat~Lal Bhatnagar, Christian Theobalt, and
  Gerard Pons-Moll.
\newblock Learning to reconstruct people in clothing from a single rgb camera.
\newblock In {\em Proceedings of the IEEE/CVF Conference on Computer Vision and
  Pattern Recognition}, pages 1175--1186, 2019.

\bibitem{alldieck2018video}
Thiemo Alldieck, Marcus Magnor, Weipeng Xu, Christian Theobalt, and Gerard
  Pons-Moll.
\newblock Video based reconstruction of 3d people models.
\newblock In {\em Proceedings of the IEEE Conference on Computer Vision and
  Pattern Recognition}, pages 8387--8397, 2018.

\bibitem{arun1987least}
K~Somani Arun, Thomas~S Huang, and Steven~D Blostein.
\newblock Least-squares fitting of two 3-d point sets.
\newblock {\em IEEE Transactions on pattern analysis and machine intelligence},
  (5):698--700, 1987.

\bibitem{bagautdinov2021driving}
Timur Bagautdinov, Chenglei Wu, Tomas Simon, Fabian Prada, Takaaki Shiratori,
  Shih-En Wei, Weipeng Xu, Yaser Sheikh, and Jason Saragih.
\newblock Driving-signal aware full-body avatars.
\newblock {\em ACM Transactions on Graphics (TOG)}, 40(4):1--17, 2021.

\bibitem{bloesch2017rovio}
Michael Bloesch, Michael Burri, Sammy Omari, Marco Hutter, and Roland Siegwart.
\newblock Iterated extended kalman filter based visual-inertial odometry using
  direct photometric feedback.
\newblock {\em The International Journal of Robotics Research},
  36(10):1053--1072, 2017.

\bibitem{bogo2015detailed}
Federica Bogo, Michael~J Black, Matthew Loper, and Javier Romero.
\newblock Detailed full-body reconstructions of moving people from monocular
  rgb-d sequences.
\newblock In {\em Proceedings of the IEEE international conference on computer
  vision}, pages 2300--2308, 2015.

\bibitem{bogo2017dynamic}
Federica Bogo, Javier Romero, Gerard Pons-Moll, and Michael~J Black.
\newblock Dynamic faust: Registering human bodies in motion.
\newblock In {\em Proceedings of the IEEE conference on computer vision and
  pattern recognition}, pages 6233--6242, 2017.

\bibitem{Burri25012016}
Michael Burri, Janosch Nikolic, Pascal Gohl, Thomas Schneider, Joern Rehder,
  Sammy Omari, Markus~W Achtelik, and Roland Siegwart.
\newblock The euroc micro aerial vehicle datasets.
\newblock {\em The International Journal of Robotics Research}, 2016.

\bibitem{campos2021orb}
Carlos Campos, Richard Elvira, Juan J~G{\'o}mez Rodr{\'\i}guez, Jos{\'e}~MM
  Montiel, and Juan~D Tard{\'o}s.
\newblock {ORB-SLAM3}: An accurate open-source library for visual,
  visual--inertial, and multimap slam.
\newblock {\em IEEE Transactions on Robotics}, 2021.

\bibitem{carlevaris2016university}
Nicholas Carlevaris-Bianco, Arash~K Ushani, and Ryan~M Eustice.
\newblock University of michigan north campus long-term vision and lidar
  dataset.
\newblock {\em The International Journal of Robotics Research},
  35(9):1023--1035, 2016.

\bibitem{carranza2003free}
Joel Carranza, Christian Theobalt, Marcus~A Magnor, and Hans-Peter Seidel.
\newblock Free-viewpoint video of human actors.
\newblock {\em ACM transactions on graphics (TOG)}, 22(3):569--577, 2003.

\bibitem{ceriani2009rawseeds}
Simone Ceriani, Giulio Fontana, Alessandro Giusti, Daniele Marzorati, Matteo
  Matteucci, Davide Migliore, Davide Rizzi, Domenico~G Sorrenti, and Pierluigi
  Taddei.
\newblock Rawseeds ground truth collection systems for indoor self-localization
  and mapping.
\newblock {\em Autonomous Robots}, 27(4):353--371, 2009.

\bibitem{collet2015high}
Alvaro Collet, Ming Chuang, Pat Sweeney, Don Gillett, Dennis Evseev, David
  Calabrese, Hugues Hoppe, Adam Kirk, and Steve Sullivan.
\newblock High-quality streamable free-viewpoint video.
\newblock {\em ACM Transactions on Graphics (ToG)}, 34(4):1--13, 2015.

\bibitem{Cortes_2018_ECCV}
Santiago Cortes, Arno Solin, Esa Rahtu, and Juho Kannala.
\newblock Advio: An authentic dataset for visual-inertial odometry.
\newblock In {\em Proceedings of the European Conference on Computer Vision
  (ECCV)}, September 2018.

\bibitem{Delmerico2019drone}
Jeffrey Delmerico, Titus Cieslewski, Henri Rebecq, Matthias Faessler, and
  Davide Scaramuzza.
\newblock Are we ready for autonomous drone racing? the uzh-fpv drone racing
  dataset.
\newblock In {\em 2019 International Conference on Robotics and Automation
  (ICRA)}, pages 6713--6719, 2019.

\bibitem{delmerico2018benchmark}
Jeffrey Delmerico and Davide Scaramuzza.
\newblock A benchmark comparison of monocular visual-inertial odometry
  algorithms for flying robots.
\newblock In {\em 2018 IEEE International Conference on Robotics and Automation
  (ICRA)}, pages 2502--2509. IEEE, 2018.

\bibitem{dou2016fusion4d}
Mingsong Dou, Sameh Khamis, Yury Degtyarev, Philip Davidson, Sean~Ryan Fanello,
  Adarsh Kowdle, Sergio~Orts Escolano, Christoph Rhemann, David Kim, Jonathan
  Taylor, et~al.
\newblock Fusion4d: Real-time performance capture of challenging scenes.
\newblock {\em ACM Transactions on Graphics (ToG)}, 35(4):1--13, 2016.

\bibitem{DBLP:journals/corr/EngelKC16}
Jakob Engel, Vladlen Koltun, and Daniel Cremers.
\newblock Direct sparse odometry.
\newblock {\em CoRR}, abs/1607.02565, 2016.

\bibitem{faizullin2021twist}
Marsel Faizullin, Anastasiia Kornilova, Azat Akhmetyanov, and Gonzalo Ferrer.
\newblock Twist-n-sync: Software clock synchronization with microseconds
  accuracy using mems-gyroscopes.
\newblock {\em Sensors}, 21(1):68, 2021.

\bibitem{faizullin2021synchronized}
Marsel Faizullin, Anastasiia Kornilova, Azat Akhmetyanov, Konstantin Pakulev,
  Andrey Sadkov, and Gonzalo Ferrer.
\newblock Synchronized smartphone video recording system of depth and rgb image
  frames with sub-millisecond precision, 2021.

\bibitem{Forster2014ICRA}
Christian Forster, Matia Pizzoli, and Davide Scaramuzza.
\newblock {SVO}: Fast semi-direct monocular visual odometry.
\newblock In {\em IEEE International Conference on Robotics and Automation
  (ICRA)}, 2014.

\bibitem{7782863}
Christian Forster, Zichao Zhang, Michael Gassner, Manuel Werlberger, and Davide
  Scaramuzza.
\newblock Svo: Semidirect visual odometry for monocular and multicamera
  systems.
\newblock {\em IEEE Transactions on Robotics}, 33(2):249--265, 2017.

\bibitem{gafni2021dynamic}
Guy Gafni, Justus Thies, Michael Zollhofer, and Matthias Nie{\ss}ner.
\newblock Dynamic neural radiance fields for monocular 4d facial avatar
  reconstruction.
\newblock In {\em Proceedings of the IEEE/CVF Conference on Computer Vision and
  Pattern Recognition}, pages 8649--8658, 2021.

\bibitem{gao2020portrait}
Chen Gao, Yichang Shih, Wei-Sheng Lai, Chia-Kai Liang, and Jia-Bin Huang.
\newblock Portrait neural radiance fields from a single image.
\newblock {\em arXiv preprint arXiv:2012.05903}, 2020.

\bibitem{gao2018ldso}
Xiang Gao, Rui Wang, Nikolaus Demmel, and Daniel Cremers.
\newblock Ldso: Direct sparse odometry with loop closure.
\newblock In {\em 2018 IEEE/RSJ International Conference on Intelligent Robots
  and Systems (IROS)}, pages 2198--2204. IEEE, 2018.

\bibitem{geiger2013vision}
Andreas Geiger, Philip Lenz, Christoph Stiller, and Raquel Urtasun.
\newblock Vision meets robotics: The kitti dataset.
\newblock {\em The International Journal of Robotics Research},
  32(11):1231--1237, 2013.

\bibitem{Geneva2020ICRA}
Patrick Geneva, Kevin Eckenhoff, Woosik Lee, Yulin Yang, and Guoquan Huang.
\newblock {OpenVINS}: A research platform for visual-inertial estimation.
\newblock In {\em Proc. of the IEEE International Conference on Robotics and
  Automation}, Paris, France, 2020.

\bibitem{guo2019relightables}
Kaiwen Guo, Peter Lincoln, Philip Davidson, Jay Busch, Xueming Yu, Matt Whalen,
  Geoff Harvey, Sergio Orts-Escolano, Rohit Pandey, Jason Dourgarian, et~al.
\newblock The relightables: Volumetric performance capture of humans with
  realistic relighting.
\newblock {\em ACM Transactions on Graphics (TOG)}, 38(6):1--19, 2019.

\bibitem{guo2017real}
Kaiwen Guo, Feng Xu, Tao Yu, Xiaoyang Liu, Qionghai Dai, and Yebin Liu.
\newblock Real-time geometry, albedo, and motion reconstruction using a single
  rgb-d camera.
\newblock {\em ACM Transactions on Graphics (ToG)}, 36(4):1, 2017.

\bibitem{hong2018dynamic}
Sungjin Hong and Yejin Kim.
\newblock Dynamic pose estimation using multiple rgb-d cameras.
\newblock {\em Sensors}, 18(11):3865, 2018.

\bibitem{ionescu2013human3}
Catalin Ionescu, Dragos Papava, Vlad Olaru, and Cristian Sminchisescu.
\newblock Human3. 6m: Large scale datasets and predictive methods for 3d human
  sensing in natural environments.
\newblock {\em IEEE transactions on pattern analysis and machine intelligence},
  36(7):1325--1339, 2013.

\bibitem{jafarian2021learning}
Yasamin Jafarian and Hyun~Soo Park.
\newblock Learning high fidelity depths of dressed humans by watching social
  media dance videos.
\newblock In {\em Proceedings of the IEEE/CVF Conference on Computer Vision and
  Pattern Recognition}, pages 12753--12762, 2021.

\bibitem{jeon2021run}
Jinwoo Jeon, Sungwook Jung, Eungchang Lee, Duckyu Choi, and Hyun Myung.
\newblock Run your visual-inertial odometry on nvidia jetson: Benchmark tests
  on a micro aerial vehicle.
\newblock {\em IEEE Robotics and Automation Letters}, 6(3):5332--5339, 2021.

\bibitem{joo2017panoptic}
Hanbyul Joo, Tomas Simon, Xulong Li, Hao Liu, Lei Tan, Lin Gui, Sean Banerjee,
  Timothy~Scott Godisart, Bart Nabbe, Iain Matthews, Takeo Kanade, Shohei
  Nobuhara, and Yaser Sheikh.
\newblock Panoptic studio: A massively multiview system for social interaction
  capture.
\newblock {\em IEEE Transactions on Pattern Analysis and Machine Intelligence},
  2017.

\bibitem{joo2018total}
Hanbyul Joo, Tomas Simon, and Yaser Sheikh.
\newblock Total capture: A 3d deformation model for tracking faces, hands, and
  bodies.
\newblock In {\em Proceedings of the IEEE conference on computer vision and
  pattern recognition}, pages 8320--8329, 2018.

\bibitem{kornilova2021mom}
Anastasiia Kornilova and Gonzalo Ferrer.
\newblock Be your own benchmark: No-reference trajectory metric on registered
  point clouds.
\newblock In {\em 2021 European Conference on Mobile Robots (ECMR)}, pages
  1--8, 2021.

\bibitem{leuteneger2014okvis}
Stefan Leutenegger, Simon Lynen, Michael Bosse, Roland Siegwart, and Paul
  Furgale.
\newblock Keyframe-based visual-inertial odometry using nonlinear optimization.
\newblock {\em The International Journal of Robotics Research}, 34, 02 2014.

\bibitem{li20133d}
Hao Li, Etienne Vouga, Anton Gudym, Linjie Luo, Jonathan~T Barron, and Gleb
  Gusev.
\newblock 3d self-portraits.
\newblock {\em ACM Transactions on Graphics (TOG)}, 32(6):1--9, 2013.

\bibitem{PRCV-LiYHZB2019}
Jinyu Li, Bangbang Yang, Kai Huang, Guofeng Zhang, and Hujun Bao.
\newblock Robust and efficient visual-inertial odometry with multi-plane
  priors.
\newblock In {\em Pattern Recognition and Computer Vision - Second Chinese
  Conference, {PRCV} 2019, Xi'an, China, November 8-11, 2019, Proceedings, Part
  {III}}, volume 11859 of {\em Lecture Notes in Computer Science}, pages
  283--295. Springer, 2019.

\bibitem{li2019learning}
Zhengqi Li, Tali Dekel, Forrester Cole, Richard Tucker, Noah Snavely, Ce Liu,
  and William~T Freeman.
\newblock Learning the depths of moving people by watching frozen people.
\newblock In {\em Proceedings of the IEEE/CVF Conference on Computer Vision and
  Pattern Recognition}, pages 4521--4530, 2019.

\bibitem{lin2021barf}
Chen-Hsuan Lin, Wei-Chiu Ma, Antonio Torralba, and Simon Lucey.
\newblock {BARF}: Bundle-adjusting neural radiance fields.
\newblock {\em arXiv preprint arXiv:2104.06405}, 2021.

\bibitem{lombardi2019neuralvolumes}
Stephen Lombardi, Tomas Simon, Jason Saragih, Gabriel Schwartz, Andreas
  Lehrmann, and Yaser Sheikh.
\newblock Neural volumes: Learning dynamic renderable volumes from images.
\newblock {\em ACM Trans. Graph.}, 38(4), July 2019.

\bibitem{loper2015smpl}
Matthew Loper, Naureen Mahmood, Javier Romero, Gerard Pons-Moll, and Michael~J
  Black.
\newblock Smpl: A skinned multi-person linear model.
\newblock {\em ACM transactions on graphics (TOG)}, 34(6):1--16, 2015.

\bibitem{mescheder2019occupancy}
Lars Mescheder, Michael Oechsle, Michael Niemeyer, Sebastian Nowozin, and
  Andreas Geiger.
\newblock Occupancy networks: Learning 3d reconstruction in function space.
\newblock In {\em Proceedings of the IEEE/CVF Conference on Computer Vision and
  Pattern Recognition}, pages 4460--4470, 2019.

\bibitem{mildenhall2020nerf}
Ben Mildenhall, Pratul~P Srinivasan, Matthew Tancik, Jonathan~T Barron, Ravi
  Ramamoorthi, and Ren Ng.
\newblock Nerf: Representing scenes as neural radiance fields for view
  synthesis.
\newblock In {\em European conference on computer vision}, pages 405--421.
  Springer, 2020.

\bibitem{4209642}
Anastasios~I. Mourikis and Stergios~I. Roumeliotis.
\newblock A multi-state constraint kalman filter for vision-aided inertial
  navigation.
\newblock In {\em Proceedings 2007 IEEE International Conference on Robotics
  and Automation}, pages 3565--3572, 2007.

\bibitem{mur2015orb}
Raul Mur-Artal, Jose Maria~Martinez Montiel, and Juan~D Tardos.
\newblock {ORB-SLAM: a versatile and accurate monocular SLAM system}.
\newblock {\em IEEE transactions on robotics}, 31(5):1147--1163, 2015.

\bibitem{mur2017orb}
Raul Mur-Artal and Juan~D Tard{\'o}s.
\newblock {ORB-SLAM2}: An open-source slam system for monocular, stereo, and
  rgb-d cameras.
\newblock {\em IEEE transactions on robotics}, 33(5):1255--1262, 2017.

\bibitem{nikolic2016non}
Janosch Nikolic, Michael Burri, Igor Gilitschenski, Juan Nieto, and Roland
  Siegwart.
\newblock Non-parametric extrinsic and intrinsic calibration of visual-inertial
  sensor systems.
\newblock {\em IEEE Sensors Journal}, 16(13):5433--5443, 2016.

\bibitem{olson2011apriltag}
Edwin Olson.
\newblock Apriltag: A robust and flexible visual fiducial system.
\newblock In {\em 2011 IEEE International Conference on Robotics and
  Automation}, pages 3400--3407. IEEE, 2011.

\bibitem{omran2018neural}
Mohamed Omran, Christoph Lassner, Gerard Pons-Moll, Peter Gehler, and Bernt
  Schiele.
\newblock Neural body fitting: Unifying deep learning and model based human
  pose and shape estimation.
\newblock In {\em 2018 international conference on 3D vision (3DV)}, pages
  484--494. IEEE, 2018.

\bibitem{park2019deepsdf}
Jeong~Joon Park, Peter Florence, Julian Straub, Richard Newcombe, and Steven
  Lovegrove.
\newblock Deepsdf: Learning continuous signed distance functions for shape
  representation.
\newblock In {\em Proceedings of the IEEE/CVF Conference on Computer Vision and
  Pattern Recognition}, pages 165--174, 2019.

\bibitem{park2021nerfies}
Keunhong Park, Utkarsh Sinha, Jonathan~T Barron, Sofien Bouaziz, Dan~B Goldman,
  Steven~M Seitz, and Ricardo Martin-Brualla.
\newblock Nerfies: Deformable neural radiance fields.
\newblock In {\em Proceedings of the IEEE/CVF International Conference on
  Computer Vision}, pages 5865--5874, 2021.

\bibitem{penncos}
Bernd Pfrommer, Nitin Sanket, Kostas Daniilidis, and Jonas Cleveland.
\newblock Penncosyvio: A challenging visual inertial odometry benchmark.
\newblock In {\em 2017 IEEE International Conference on Robotics and Automation
  (ICRA)}, pages 3847--3854, 2017.

\bibitem{pumarola20193dpeople}
Albert Pumarola, Jordi Sanchez-Riera, Gary Choi, Alberto Sanfeliu, and Francesc
  Moreno-Noguer.
\newblock 3dpeople: Modeling the geometry of dressed humans.
\newblock In {\em Proceedings of the IEEE/CVF international conference on
  computer vision}, pages 2242--2251, 2019.

\bibitem{qin2019global}
Tong Qin, Shaozu Cao, Jie Pan, and Shaojie Shen.
\newblock A general optimization-based framework for global pose estimation
  with multiple sensors, 2019.

\bibitem{qin2017vins}
Tong Qin, Peiliang Li, and Shaojie Shen.
\newblock Vins-mono: A robust and versatile monocular visual-inertial state
  estimator.
\newblock {\em IEEE Transactions on Robotics}, 34(4):1004--1020, 2018.

\bibitem{qin2019local}
Tong Qin, Jie Pan, Shaozu Cao, and Shaojie Shen.
\newblock A general optimization-based framework for local odometry estimation
  with multiple sensors, 2019.

\bibitem{qin2020u2}
Xuebin Qin, Zichen Zhang, Chenyang Huang, Masood Dehghan, Osmar~R Zaiane, and
  Martin Jagersand.
\newblock U2-net: Going deeper with nested u-structure for salient object
  detection.
\newblock {\em Pattern Recognition}, 106:107404, 2020.

\bibitem{razlaw2015evaluation}
Jan Razlaw, David Droeschel, Dirk Holz, and Sven Behnke.
\newblock Evaluation of registration methods for sparse 3d laser scans.
\newblock In {\em 2015 European Conference on Mobile Robots (ECMR)}, pages
  1--7. IEEE, 2015.

\bibitem{rehder2016extending}
Joern Rehder, Janosch Nikolic, Thomas Schneider, Timo Hinzmann, and Roland
  Siegwart.
\newblock Extending kalibr: Calibrating the extrinsics of multiple imus and of
  individual axes.
\newblock In {\em 2016 IEEE International Conference on Robotics and Automation
  (ICRA)}, pages 4304--4311. IEEE, 2016.

\bibitem{Riegler2020FVS}
Gernot Riegler and Vladlen Koltun.
\newblock Free view synthesis.
\newblock In {\em European Conference on Computer Vision}, 2020.

\bibitem{riegler2021stable}
Gernot Riegler and Vladlen Koltun.
\newblock Stable view synthesis.
\newblock In {\em Proceedings of the IEEE/CVF Conference on Computer Vision and
  Pattern Recognition}, pages 12216--12225, 2021.

\bibitem{Rosinol20icra-Kimera}
Antoni Rosinol, Marcus Abate, Yun Chang, and Luca Carlone.
\newblock Kimera: an open-source library for real-time metric-semantic
  localization and mapping.
\newblock In {\em IEEE Intl. Conf. on Robotics and Automation (ICRA)}, 2020.

\bibitem{saito2019pifu}
Shunsuke Saito, Zeng Huang, Ryota Natsume, Shigeo Morishima, Angjoo Kanazawa,
  and Hao Li.
\newblock Pifu: Pixel-aligned implicit function for high-resolution clothed
  human digitization.
\newblock In {\em Proceedings of the IEEE/CVF International Conference on
  Computer Vision}, pages 2304--2314, 2019.

\bibitem{saito2020pifuhd}
Shunsuke Saito, Tomas Simon, Jason Saragih, and Hanbyul Joo.
\newblock Pifuhd: Multi-level pixel-aligned implicit function for
  high-resolution 3d human digitization.
\newblock In {\em Proceedings of the IEEE/CVF Conference on Computer Vision and
  Pattern Recognition}, pages 84--93, 2020.

\bibitem{schoenberger2016sfm}
Johannes~Lutz Sch\"{o}nberger and Jan-Michael Frahm.
\newblock Structure-from-motion revisited.
\newblock In {\em Conference on Computer Vision and Pattern Recognition
  (CVPR)}, 2016.

\bibitem{schoenberger2016mvs}
Johannes~Lutz Sch\"{o}nberger, Enliang Zheng, Marc Pollefeys, and Jan-Michael
  Frahm.
\newblock Pixelwise view selection for unstructured multi-view stereo.
\newblock In {\em European Conference on Computer Vision (ECCV)}, 2016.

\bibitem{tumvi}
David Schubert, Thore Goll, Nikolaus Demmel, Vladyslav Usenko, Jörg Stückler,
  and Daniel Cremers.
\newblock The tum vi benchmark for evaluating visual-inertial odometry.
\newblock In {\em 2018 IEEE/RSJ International Conference on Intelligent Robots
  and Systems (IROS)}, pages 1680--1687, 2018.

\bibitem{sevastopolsky2020relightable}
Artem Sevastopolsky, Savva Ignatiev, Gonzalo Ferrer, Evgeny Burnaev, and Victor
  Lempitsky.
\newblock Relightable 3d head portraits from a smartphone video.
\newblock {\em arXiv preprint arXiv:2012.09963}, 2020.

\bibitem{shapiro2014rapid}
Ari Shapiro, Andrew Feng, Ruizhe Wang, Hao Li, Mark Bolas, Gerard Medioni, and
  Evan Suma.
\newblock Rapid avatar capture and simulation using commodity depth sensors.
\newblock {\em Computer Animation and Virtual Worlds}, 25(3-4):201--211, 2014.

\bibitem{shysheya2019textured}
Aliaksandra Shysheya, Egor Zakharov, Kara-Ali Aliev, Renat Bashirov, Egor
  Burkov, Karim Iskakov, Aleksei Ivakhnenko, Yury Malkov, Igor Pasechnik,
  Dmitry Ulyanov, et~al.
\newblock Textured neural avatars.
\newblock In {\em Proceedings of the IEEE/CVF Conference on Computer Vision and
  Pattern Recognition}, pages 2387--2397, 2019.

\bibitem{sigal2010humaneva}
Leonid Sigal, Alexandru~O Balan, and Michael~J Black.
\newblock Humaneva: Synchronized video and motion capture dataset and baseline
  algorithm for evaluation of articulated human motion.
\newblock {\em International journal of computer vision}, 87(1-2):4, 2010.

\bibitem{lsm6dso}
STMicroelectronics.
\newblock {\em {iNEMO} inertial module: always-on 3D accelerometer and 3D
  gyroscope}, 2019.
\newblock Rev. 2.

\bibitem{sturm12iros}
J. Sturm, N. Engelhard, F. Endres, W. Burgard, and D. Cremers.
\newblock A benchmark for the evaluation of rgb-d slam systems.
\newblock In {\em Proc. of the International Conference on Intelligent Robot
  Systems (IROS)}, Oct. 2012.

\bibitem{openvslam2019}
Shinya Sumikura, Mikiya Shibuya, and Ken Sakurada.
\newblock {OpenVSLAM: A Versatile Visual SLAM Framework}.
\newblock In {\em Proceedings of the 27th ACM International Conference on
  Multimedia}, MM '19, pages 2292--2295, New York, NY, USA, 2019. ACM.

\bibitem{tiwari2020sizer}
Garvita Tiwari, Bharat~Lal Bhatnagar, Tony Tung, and Gerard Pons-Moll.
\newblock Sizer: A dataset and model for parsing 3d clothing and learning size
  sensitive 3d clothing.
\newblock In {\em European Conference on Computer Vision}, pages 1--18.
  Springer, 2020.

\bibitem{toderici2013uhdb11}
George Toderici, Georgios Evangelopoulos, Tianhong Fang, Theoharis Theoharis,
  and Ioannis~A Kakadiaris.
\newblock Uhdb11 database for 3d-2d face recognition.
\newblock In {\em Pacific-Rim Symposium on Image and Video Technology}, pages
  73--86. Springer, 2013.

\bibitem{tsuchida2019aist}
Shuhei Tsuchida, Satoru Fukayama, Masahiro Hamasaki, and Masataka Goto.
\newblock {AIST} dance video database: Multi-genre, multi-dancer, and
  multi-camera database for dance information processing.
\newblock In {\em ISMIR}, pages 501--510, 2019.

\bibitem{umeyama1991}
Shinji Umeyama.
\newblock Least-squares estimation of transformation parameters between two
  point patterns.
\newblock {\em IEEE Transactions on pattern analysis and machine intelligence},
  13(4):376--380, 1991.

\bibitem{vlasic2009dynamic}
Daniel Vlasic, Pieter Peers, Ilya Baran, Paul Debevec, Jovan Popovi{\'c},
  Szymon Rusinkiewicz, and Wojciech Matusik.
\newblock Dynamic shape capture using multi-view photometric stereo.
\newblock In {\em ACM SIGGRAPH Asia 2009 papers}, pages 1--11. 2009.

\bibitem{wang2019near}
Cheng Wang, Yu Zhao, Jiabin Guo, Ling Pei, Yue Wang, and Haiwei Liu.
\newblock {NEAR: The NetEase AR} oriented visual inertial dataset.
\newblock In {\em International Symposium on Mixed and Augmented Reality
  Adjunct (ISMAR-Adjunct)}, pages 366--371. IEEE, 2019.

\bibitem{wang2021learning}
Ziyan Wang, Timur Bagautdinov, Stephen Lombardi, Tomas Simon, Jason Saragih,
  Jessica Hodgins, and Michael Zollhofer.
\newblock Learning compositional radiance fields of dynamic human heads.
\newblock In {\em Proceedings of the IEEE/CVF Conference on Computer Vision and
  Pattern Recognition}, pages 5704--5713, 2021.

\bibitem{xu2020planar}
Qingshan Xu and Wenbing Tao.
\newblock Planar prior assisted patchmatch multi-view stereo.
\newblock In {\em Proceedings of the AAAI Conference on Artificial
  Intelligence}, volume~34, pages 12516--12523, 2020.

\bibitem{xu2018monoperfcap}
Weipeng Xu, Avishek Chatterjee, Michael Zollh{\"o}fer, Helge Rhodin, Dushyant
  Mehta, Hans-Peter Seidel, and Christian Theobalt.
\newblock Monoperfcap: Human performance capture from monocular video.
\newblock {\em ACM Transactions on Graphics (ToG)}, 37(2):1--15, 2018.

\bibitem{Jinlong2016inria}
Jinlong Yang, Jean-S{\'e}bastien Franco, Franck H{\'e}troy-Wheeler, and
  Stefanie Wuhrer.
\newblock Estimation of human body shape in motion with wide clothing.
\newblock In Bastian Leibe, Jiri Matas, Nicu Sebe, and Max Welling, editors,
  {\em Computer Vision -- ECCV 2016}, pages 439--454, Cham, 2016. Springer
  International Publishing.

\bibitem{yu2018doublefusion}
Tao Yu, Zerong Zheng, Kaiwen Guo, Jianhui Zhao, Qionghai Dai, Hao Li, Gerard
  Pons-Moll, and Yebin Liu.
\newblock Doublefusion: Real-time capture of human performances with inner body
  shapes from a single depth sensor.
\newblock In {\em Proceedings of the IEEE conference on computer vision and
  pattern recognition}, pages 7287--7296, 2018.

\bibitem{yu2020humbi}
Zhixuan Yu, Jae~Shin Yoon, In~Kyu Lee, Prashanth Venkatesh, Jaesik Park, Jihun
  Yu, and Hyun~Soo Park.
\newblock Humbi: A large multiview dataset of human body expressions.
\newblock In {\em Proceedings of the IEEE/CVF Conference on Computer Vision and
  Pattern Recognition}, pages 2990--3000, 2020.

\bibitem{zakharov2020fast}
Egor Zakharov, Aleksei Ivakhnenko, Aliaksandra Shysheya, and Victor Lempitsky.
\newblock Fast bi-layer neural synthesis of one-shot realistic head avatars.
\newblock In {\em European Conference on Computer Vision}, pages 524--540.
  Springer, 2020.

\bibitem{Zakharov2019talking}
Egor Zakharov, Aliaksandra Shysheya, Egor Burkov, and Victor Lempitsky.
\newblock Few-shot adversarial learning of realistic neural talking head
  models.
\newblock In {\em Proceedings of the IEEE/CVF International Conference on
  Computer Vision (ICCV)}, October 2019.

\bibitem{zhang2017detailed}
Chao Zhang, Sergi Pujades, Michael~J Black, and Gerard Pons-Moll.
\newblock Detailed, accurate, human shape estimation from clothed 3d scan
  sequences.
\newblock In {\em Proceedings of the IEEE Conference on Computer Vision and
  Pattern Recognition}, pages 4191--4200, 2017.

\bibitem{zhang2012real}
Licong Zhang, J{\"u}rgen Sturm, Daniel Cremers, and Dongheui Lee.
\newblock Real-time human motion tracking using multiple depth cameras.
\newblock In {\em 2012 IEEE/RSJ International Conference on Intelligent Robots
  and Systems}, pages 2389--2395. IEEE, 2012.

\bibitem{zhang2021survey}
Shishun Zhang, Longyu Zheng, and Wenbing Tao.
\newblock Survey and evaluation of rgb-d slam.
\newblock {\em IEEE Access}, 9:21367--21387, 2021.

\bibitem{zhang2021lightweight}
Yuxiang Zhang, Zhe Li, Liang An, Mengcheng Li, Tao Yu, and Yebin Liu.
\newblock Lightweight multi-person total motion capture using sparse multi-view
  cameras.
\newblock In {\em Proceedings of the IEEE/CVF International Conference on
  Computer Vision}, pages 5560--5569, 2021.

\bibitem{zheng2021deepmulticap}
Yang Zheng, Ruizhi Shao, Yuxiang Zhang, Tao Yu, Zerong Zheng, Qionghai Dai, and
  Yebin Liu.
\newblock Deepmulticap: Performance capture of multiple characters using sparse
  multiview cameras.
\newblock {\em arXiv preprint arXiv:2105.00261}, 2021.

\bibitem{zuniga2020vi}
David Zu{\~n}iga-No{\"e}l, Alberto Jaenal, Ruben Gomez-Ojeda, and Javier
  Gonzalez-Jimenez.
\newblock The uma-vi dataset: Visual--inertial odometry in low-textured and
  dynamic illumination environments.
\newblock {\em The International Journal of Robotics Research},
  39(9):1052--1060, 2020.

\end{thebibliography}
}

\end{document}